\documentclass{article}

\usepackage[preprint]{neurips_2026}

\usepackage[utf8]{inputenc}
\usepackage[T1]{fontenc}
\usepackage{hyperref}
\usepackage{url}
\usepackage{booktabs}
\usepackage{amsfonts}
\usepackage{amsmath}
\usepackage{amssymb}
\usepackage{nicefrac}
\usepackage{microtype}
\usepackage{xcolor}
\usepackage{graphicx}
\usepackage{multirow}
\usepackage{array}
\usepackage{placeins}
\usepackage{tikz}
\usetikzlibrary{arrows.meta,positioning}

\graphicspath{{figures/}}

\newcommand{\relAblock}{\ensuremath{r_{A/\mathrm{blk}}}}
\newcommand{\relAres}{\ensuremath{r_{A/\mathrm{res}}}}
\newcommand{\relArouted}{\ensuremath{r_{A/\mathrm{rt}}}}
\newcommand{\relBres}{\ensuremath{r_{B/\mathrm{res}}}}
\newcommand{\blockres}{\ensuremath{r_{\mathrm{blk}/\mathrm{res}}}}
\newcommand{\Sg}{\ensuremath{\Sigma g}}

\title{Routing Divergence Is Not Evidence of Behavioral Influence in Same-Weight MoE Self-Distillation}

\author{%
  Cedric Caruzzo \\
  Lunit \\
  \texttt{cedric.caruzzo@lunit.io}
  \And
  Donggeun Yoo \\
  Lunit \\
  \texttt{dgyoo@lunit.io}
  \And
  Tae Soo Kim\thanks{Corresponding author.} \\
  Lunit \\
  \texttt{taesoo.kim@lunit.io}
}

\begin{document}
\maketitle

\begin{abstract}
Two Mixture-of-Experts (MoE) forward passes can share every weight yet route the same token through
different experts. This creates a possible blind spot in same-weight self-distillation, where a
demonstration-conditioned teacher supervises a query-only student. We study this mismatch in its
single-step form, with frozen weights rather than as a proxy for a full training trajectory. An exact
blockwise decomposition separates a routing term, which changes gates at fixed content, from a
dense-like content term. Across seven open-weight checkpoints and two domains, the routing term spans
only $1.6\times$ as a fraction of block output, while its residual-stream exposure spans
$3.2\times$. Exposure is ordered by the routed block's share of the residual. Scaling the always-on
backbone in two confirmatory models moves exposure
monotonically; common-mode controls support a mass-and-coherence mechanism rather than denominator
dilution alone. Preregistered PubMedQA patches on three models show that the full routing term moves outputs
by less than half the natural context effect and is largely reproduced by matched-norm noise, whereas
the content term is strongly direction-specific. Scale and merged-expert probes show that the narrow
block-level range is not universal, although exposure remains small at the tested boundaries. Router
movement alone is therefore not evidence of behavioral influence: measure exposure first, and use a
behavioral intervention when the decision matters.
\end{abstract}

\section{Introduction}
\label{sec:intro}
Post-training a Mixture-of-Experts model introduces a complication absent from dense models: two
forward passes can share every parameter yet execute different sparse computations. In same-weight
self-distillation, a demonstration-conditioned teacher and a query-only student can route the same
token to different experts. Existing work measures or suppresses router drift
\citep{dai2022stablemoe, ma2025r3, xie2026same}, but gate divergence alone does not say how much of
the induced perturbation reaches the residual stream or the model output.

We ask that downstream question in the single-step regime the method actually constructs. We hold
weights fixed and compare the teacher and student passes before any update. An exact blockwise
decomposition separates the change caused by moving the gates at fixed content from the dense-like
change caused by shifting content at fixed gates. This lets us trace routing mismatch from the routed
block to the residual stream and, through causal patches, to output behavior. Drift accumulated over
many updates, or mismatch between separately parameterized teachers and students, is outside this
study.

The natural prediction is that a router that moves more creates more downstream risk. Our
measurements invert it. Across seven checkpoints and two domains, the routing term is relatively
stable as a share of block output, but its exposure to later layers varies with the routed block's
share of the residual. The most fine-grained router is among the least exposed. Backbone rescaling on
two confirmatory checkpoints supports the causal account: always-on mass contains the mismatch
chiefly by maintaining routing coherence. Output patches then show that the routing term is bounded
and mostly magnitude-generic, while the dense-like content term is direction-specific.

The claim is deliberately narrow: routing can matter, but router movement is not itself a measure of
downstream influence. Our contributions are:
\begin{itemize}
  \item An exact $A/B$ decomposition and reconstruction gate that distinguish routing mismatch from
        content shift, together with residual exposure as the downstream quantity of interest.
  \item A seven-checkpoint, two-domain measurement showing that exposure follows backbone share, and
        a two-model intervention showing that always-on mass preserves routing coherence.
  \item Three-model causal $A/B$ patches that connect the decomposition to output behavior, plus
        scale and merged-expert probes that delimit rather than universalize the observed range.
\end{itemize}

\paragraph{Relation to companion work.} This report asks how much same-weight,
conditioning-induced routing mismatch survives from an MoE block into the residual stream and what
effect that term has when patched causally. A companion study analyzes how moved gate mass aligns
with expert-output directions; a related challenge formulation asks how to allocate a limited budget
of behavioral interventions. The works share an experimental program, but their primary estimands
are distinct: residual exposure here, correspondence geometry there, and audit allocation in the
challenge.

\section{Preliminaries and the decomposition}
\label{sec:decomp}

\begin{figure}[t]
\centering
\resizebox{\linewidth}{!}{%
\begin{tikzpicture}[font=\small,
  box/.style={draw, rounded corners=2pt, align=center, inner sep=4pt},
  blk/.style={draw, rounded corners=2pt, align=center, fill=black!5, inner sep=5pt},
  hl/.style={draw, rounded corners=2pt, align=center, fill=green!10, inner sep=4pt},
  ar/.style={-{Latex[length=2mm]}, semithick}]
\node[box] (teach) {teacher pass\\[-2pt]{\footnotesize demonstration $+$ query}};
\node[box, below=9mm of teach] (stud) {student pass\\[-2pt]{\footnotesize query only}};
\node[blk, right=12mm of teach, yshift=-10mm] (blk) {MoE block\\[-2pt]{\footnotesize shared weights $\theta$}\\[-2pt]{\footnotesize router $\to$ experts $f_j$}};
\node[box, right=13mm of blk, yshift=10mm] (t) {$g^{T},\,h^{T}$};
\node[box, right=13mm of blk, yshift=-10mm] (s) {$g^{S},\,h^{S}$};
\node[hl, right=13mm of t] (A) {$A=\sum_j (g^{T}_j-g^{S}_j)\,f_j(h^{S})$\\[-2pt]{\footnotesize routing mismatch: same content, different gates}};
\node[box, right=13mm of s] (B) {$B=\sum_j g^{T}_j\big(f_j(h^{T})-f_j(h^{S})\big)$\\[-2pt]{\footnotesize dense-like: same gates, different content}};
\draw[ar] (teach) -- (blk);
\draw[ar] (stud) -- (blk);
\draw[ar] (blk) -- (t);
\draw[ar] (blk) -- (s);
\draw[ar] (t) -- (A);
\draw[ar] (s) -- (B);
\node[box, below=18mm of stud] (div) {routing\\[-2pt]divergence};
\node[box, right=13mm of div] (dec) {decomposition\\[-2pt]{\footnotesize $A,\;B$}};
\node[box, right=13mm of dec] (fac) {routing factor $\relAblock$\\[-2pt]$\times$ backbone share $\blockres$};
\node[hl, right=13mm of fac] (exp) {exposure\\[-2pt]$\relAres$};
\node[box, right=13mm of exp] (down) {output\\[-2pt]influence};
\draw[ar] (div) -- (dec);
\draw[ar] (dec) -- (fac);
\draw[ar] (fac) -- (exp);
\draw[ar] (exp) -- (down);
\node[font=\footnotesize, below=6mm of fac, align=center] {raw divergence does not set exposure; backbone share does};
\end{tikzpicture}}
\caption{The decomposition and its causal chain. Top: two forward passes of the same model with the
same weights, one with a demonstration in context (teacher) and one without (student), produce
different gate vectors and
hidden states at each MoE block. The change in the routed output $\Delta_{\mathrm{routed}} = A + B$
splits into a routing-mismatch term $A$ (the same content $h^{S}$ sent through the two passes'
different gates) and a dense-like remainder $B$ (the teacher routing applied to the content shift).
This paper measures $A$ and its share of the residual stream (Eq.~\ref{eq:wedge}). Bottom: the
routing factor varies less than exposure in the primary screen, while backbone share maps the
block-level perturbation into residual exposure. Neither the observed range nor exposure alone is a
behavioral guarantee.}
\label{fig:concept}
\end{figure}
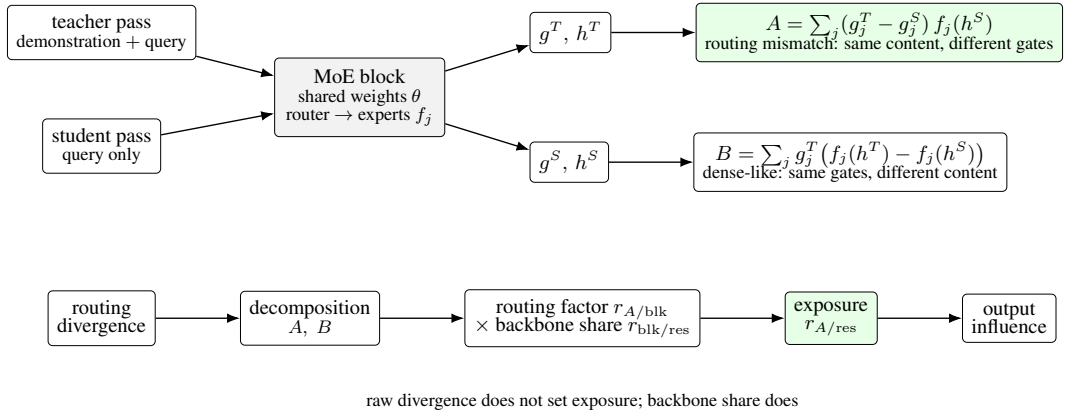

\paragraph{Setup.} We compare two forward passes of the \emph{same} model with the \emph{same}
weights (Figure~\ref{fig:concept}). In the teacher pass the input carries a demonstration (the
context a self-distillation teacher sees \citep{shenfeld2026sdft}); in the student pass it does not. Superscripts $T$ and $S$ mark the pass. At an MoE
block, for a given token, let $g_j$ be the realized gate weight on expert $j$ (the value applied to the expert output, which need not sum to one across the selected experts; see
$\Sg$ below), let $f_j(h)$ be expert $j$'s output on hidden state $h$, and let the routed output be
$\mathrm{routed} = \sum_j g_j f_j(h)$. The block adds an optional shared-expert output and returns
$y$; the residual stream carries $\mathrm{res}$ into the block.

\paragraph{Decomposition.} The change in the routed output between the two passes,
$\Delta_{\mathrm{routed}} = \mathrm{routed}^T - \mathrm{routed}^S$, splits by adding and subtracting
$\sum_j g^T_j f_j(h^S)$:
\begin{align}
  A &= \textstyle\sum_j (g^T_j - g^S_j)\, f_j(h^S), \label{eq:A}\\
  B &= \textstyle\sum_j g^T_j\, \big(f_j(h^T) - f_j(h^S)\big), \label{eq:B}\\
  \Delta_{\mathrm{routed}} &= A + B. \label{eq:AB}
\end{align}
$A$ is the routing-mismatch term: the \emph{same} hidden state $h^S$ sent through the two passes'
\emph{different} gate vectors. $B$ is the dense-like remainder: the teacher routing applied to the
content shift $f_j(h^T) - f_j(h^S)$, the term a dense model would also incur. The anchor choice
(evaluating the expert functions at $h^S$ in $A$) is one of two symmetric options; we use $h^S$
throughout and note the alternative in Appendix~\ref{app:derivation}.\footnote{The identity
\eqref{eq:AB} is exact; we verify it per model with a reconstruction gate,
$\lVert (A+B) - \Delta_{\mathrm{routed}}\rVert / \lVert\Delta_{\mathrm{routed}}\rVert < 0.02$.
Quantized expert execution does not reproduce the identity and is excluded; robustness checks are in
Appendix~\ref{app:pathologies}.}

\paragraph{Three ratios.} We report $A$'s magnitude relative to three denominators, all per token
per layer:
\begin{equation}
  \underbrace{\relAblock = \frac{\lVert A\rVert}{\lVert y\rVert}}_{\text{routing factor}},
  \qquad
  \underbrace{\blockres = \frac{\lVert y\rVert}{\lVert \mathrm{res}\rVert}}_{\text{backbone share}},
  \qquad
  \underbrace{\relAres = \frac{\lVert A\rVert}{\lVert \mathrm{res}\rVert}}_{\text{exposure}}
  = \relAblock \cdot \blockres .
  \label{eq:wedge}
\end{equation}
The routing factor is how much the routing mismatch perturbs the block's own output. The exposure
is how much it perturbs the residual stream that the rest of the network reads. Equation
\eqref{eq:wedge} (the wedge identity) says exposure is the routing factor diluted by backbone
share: the smaller the routed block's share of the residual, the more a fixed routing perturbation
is diluted. We also report $\relArouted = \lVert A\rVert / \lVert\mathrm{routed}\rVert$ (the shared
expert removed from the denominator) and the context term $\relBres = \lVert B\rVert /
\lVert\mathrm{res}\rVert$.

\begin{center}
\fbox{\begin{minipage}{0.95\linewidth}
\small
\textbf{Glossary.}\;
$A$: routing-mismatch term, same content through different gates (Eq.~\ref{eq:A}).\;
$B$: dense-like remainder, same gates on shifted content (Eq.~\ref{eq:B}).\;
\relAblock{} \emph{(routing factor)} $=\lVert A\rVert/\lVert y\rVert$, the mismatch's share of the
block output.\;
\blockres{} \emph{(backbone share)} $=\lVert y\rVert/\lVert\mathrm{res}\rVert$, the routed block's
share of the residual.\;
\relAres{} \emph{(exposure)} $=\lVert A\rVert/\lVert\mathrm{res}\rVert=\relAblock\cdot\blockres$, the
mismatch's share of the residual, the quantity that reaches the rest of the network.\;
$\Sg$: sum of applied gate weights (convention scale).
\end{minipage}}
\end{center}

\paragraph{Residual convention.} We measure exposure at two points. The \emph{screen}
(Section~\ref{sec:resultsI}) uses the pre-block residual, the input to the post-attention
layernorm. The \emph{intervention} (Section~\ref{sec:resultsII}) uses the post-block residual,
$\mathrm{res} + y$, which is the well-posed local target for scaling $y$. The two differ by the
block's own output $y$ at each token and are reported separately; every table states which is
used.

\paragraph{Gate convention and a magnitude bound.} Because $A$ uses the applied gate weights, its
raw scale depends on whether a router renormalizes, is sub-stochastic, or applies a scaled
convention. Dividing by the block output makes \relAblock{} less sensitive to that convention.
Writing $F_{\max}=\max_j\lVert f_j(h^S)\rVert$, Eq.~\ref{eq:A} gives the exact triangle bound
$\lVert A\rVert\le F_{\max}\lVert\Delta g\rVert_1$. For renormalized gate distributions this becomes
$\lVert A\rVert\le 2\Sg F_{\max}\mathrm{TV}$, making the convention scale explicit. We use the
decomposition and reconstruction gate for measurement; the bound is an envelope, not an estimator.

\paragraph{What the decomposition buys.} The split \eqref{eq:AB} is an exact identity, so it is not
a claim on its own; its value is that $A$ isolates a quantity that can be measured in isolation, the
change in the block output that comes purely from the gates moving, with hidden state and expert
weights held fixed. That is not the same as the routed output changing, which also absorbs the
content shift $B$, and it is not recoverable from a routing-distance metric such as Jaccard or TV,
which see the gate vectors but not what the experts do with them. The rest of the paper asks how
large $A$ is relative to the block and residual, what controls that exposure, and what happens when
the recorded term is injected causally at the output.

\paragraph{Scope.} Our instrument measures conditioning-induced routing mismatch between two forward
passes of the same model: teacher and student share parameters and differ only in conditioning. This
is the setting of self-distillation methods in which the teacher is the current model under
additional conditioning \citep{shenfeld2026sdft, hubotter2026sdpo}; for these methods,
conditioning-induced mismatch is not a simplification of the general fine-tuning case, it is the
whole of the routing mismatch present in the training signal. Pipelines that distill from a
separate teacher checkpoint \citep{agarwal2024gkd, qwen3_2025, glm45_2025} additionally introduce
weight-induced routing mismatch between distinct routers, which is outside the scope of the present
work.

\section{Measurement methodology}
\label{sec:method}

\paragraph{Models.} We measure seven open-weight MoE checkpoints chosen to vary the architecture axes that a containment
account should care about (Table~\ref{tab:master}): expert count and top-$k$ (8/2 for Mixtral
\citep{jiang2024mixtral} up to 128/8 for Qwen3 \citep{qwen3_2025}), presence of an always-on shared
expert (DeepSeek-V2-Lite \citep{deepseekv2}, Gravity-16B-A3B \citep{trillion2026gravity}, and
Qwen1.5-MoE-A2.7B \citep{qwen2024qwen15moe} have one; OLMoE \citep{muennighoff2024olmoe},
Phi-3.5-MoE \citep{abdin2024phi3}, and the others do not), and gate convention ($\Sg$). All seven
reproduce the decomposition identity. Larger Qwen3 and DeepSeek-V2 checkpoints enter only as
boundary probes and do not define the observed range.

\paragraph{Probes.} Two domains carry the analysis, PubMedQA \citep{jin2019pubmedqa} (medical,
$n=200$) and GSM8K \citep{cobbe2021gsm8k} (mathematical reasoning, $n=200$). Each probe pair is a question presented with and without a worked demonstration; the
demonstration is the only difference between the teacher and student inputs, so $A$ isolates the
routing response to the demonstration. Inputs use each model's chat template where present. We aggregate
per-layer-first: within a layer we sum
numerator and denominator norms across tokens and divide once, then summarize over layers. The
alternative (averaging per-token ratios) overweights small-residual tokens and inflates the
headline. We report the per-layer-first value throughout; aggregation and prompt-format checks are
in Appendix~\ref{app:pathologies}.

\paragraph{Uncertainty.} We bootstrap over the $n=200$ probe prompts (2000 resamples), recomputing
the per-layer-first ratio under each resample, to place a $95\%$ confidence interval on every
per-model headline. The intervals are tight: across the seven checkpoints and both domains, the
per-model half-widths are at or below $0.003$ for the routing factor and $0.0012$ for the exposure,
at most $5.5\%$ and $2.8\%$ of the widths of the corresponding two-domain ranges
(Figure~\ref{fig:invariance}, Appendix~\ref{app:tables}).
The intervals establish that the seven point estimates are stable, not that the underlying quantity
must cluster.

\paragraph{Routing metrics and controls.} The backbone intervention reports top-$k$ overlap (Jaccard)
and total variation (TV) on the renormalized gate distributions. Confound and format controls are
summarized in Appendix~\ref{app:pathologies}.

\begin{table}[t]
\centering
\caption{The seven primary checkpoints. Values are per-layer-first body aggregates, shown as
PubMedQA/GSM8K. Backbone share is computed directly as
$\lVert y\rVert/\lVert\mathrm{res}\rVert$ before averaging across layers; exposure uses the
pre-block residual. Prompt-bootstrap intervals are in
Appendix~\ref{app:tables}.}
\label{tab:master}
\small
\setlength{\tabcolsep}{3pt}
\resizebox{\textwidth}{!}{%
\begin{tabular}{lccccc}
\toprule
Model & experts/$k$ & shared path & \relAblock & \blockres & \relAres \\
\midrule
Qwen3-30B-A3B      & 128/8 & --          & 0.133/0.139 & 0.174/0.178 & 0.019/0.021 \\
Mixtral-8x7B       & 8/2   & --          & 0.094/0.110 & 0.271/0.283 & 0.022/0.028 \\
OLMoE-1B-7B        & 64/8  & --          & 0.124/0.122 & 0.481/0.515 & 0.056/0.061 \\
DeepSeek-V2-Lite   & 64/6  & always-on   & 0.090/0.102 & 0.356/0.364 & 0.027/0.033 \\
Gravity-16B-A3B    & 64/8  & always-on   & 0.098/0.108 & 0.470/0.511 & 0.044/0.053 \\
Qwen1.5-MoE-A2.7B & 60/4  & input-gated & 0.085/0.102 & 0.464/0.500 & 0.036/0.048 \\
Phi-3.5-MoE        & 16/2  & --          & 0.106/0.110 & 0.246/0.262 & 0.023/0.027 \\
\midrule
range & & & 0.085--0.139 & 0.174--0.515 & 0.019--0.061 \\
\bottomrule
\end{tabular}}
\end{table}

\section{Results I: backbone share orders exposure}
\label{sec:resultsI}

\begin{figure}[t]
\centering
\includegraphics[width=0.72\linewidth]{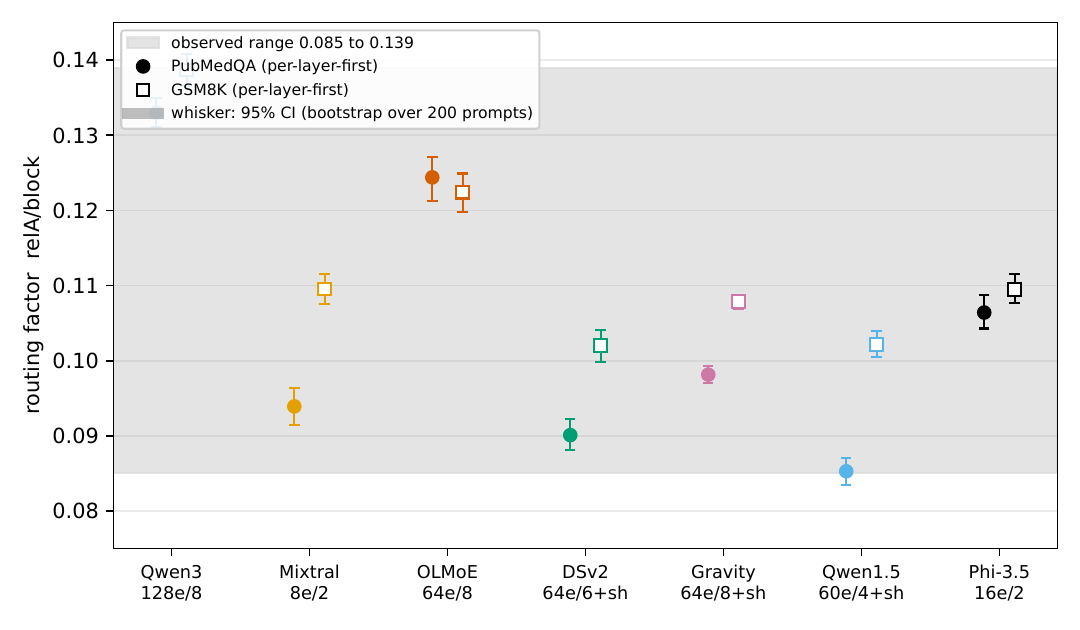}
\caption{Routing factor \relAblock{} for the seven primary checkpoints on PubMedQA (circles) and
GSM8K (squares). Whiskers are $95\%$ confidence intervals from a bootstrap over the $200$ prompts.
The precise checkpoint estimates occupy the observed $0.085$--$0.139$ range despite substantial
differences in expert count, top-$k$, shared paths, and gate conventions. The shaded range is a
description of these checkpoints, not a universal confidence band.}
\label{fig:invariance}
\end{figure}

\paragraph{The routing factor varies less than exposure.} Across the seven checkpoints and both
domains, the per-layer-first routing factor \relAblock{} lies between $0.085$ and $0.139$, a
$1.6\times$ span (Figure~\ref{fig:invariance}). Exposure \relAres{} spans $0.019$ to $0.061$,
or $3.2\times$, while directly aggregated backbone share spans $0.174$ to $0.515$. Prompt-bootstrap
intervals are narrow relative to these ranges (Appendix~\ref{app:tables}), so the checkpoint
estimates are precise. The range is nevertheless an empirical regularity of seven family-correlated
checkpoints, not a population law.

\begin{figure}[t]
\centering
\includegraphics[width=0.72\linewidth]{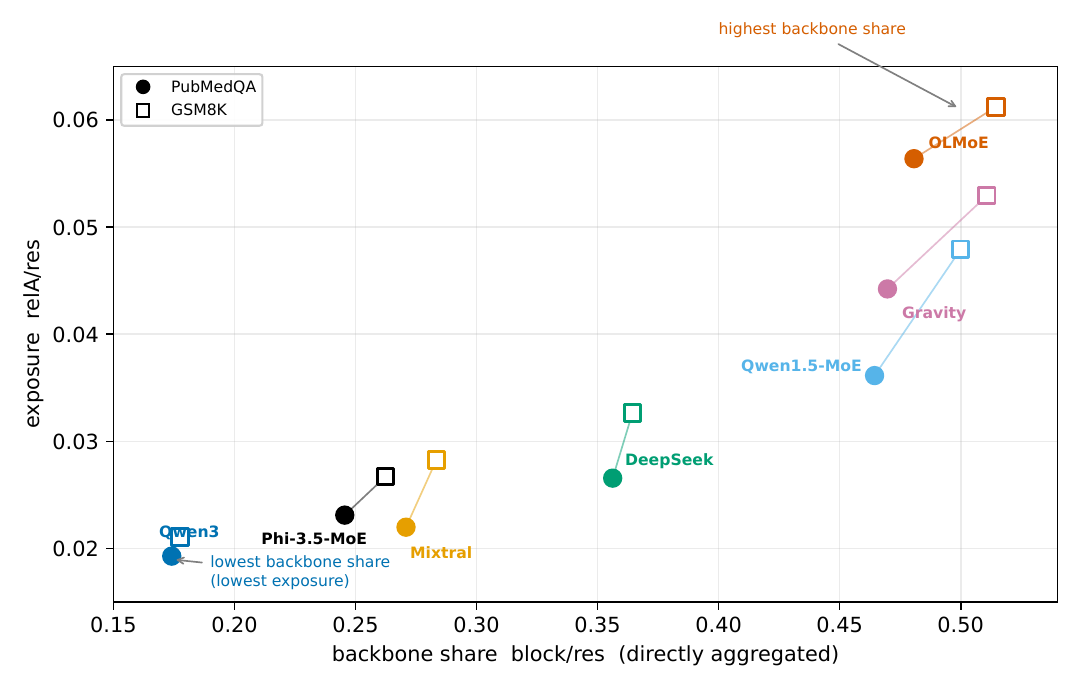}
\caption{Exposure \relAres{} against directly aggregated backbone share \blockres{} on PubMedQA
(circles) and GSM8K (squares). Each model's domain points are connected. The plot uses
$\operatorname{mean}_{\ell}(\sum\lVert y\rVert/\sum\lVert\mathrm{res}\rVert)$ rather than deriving
the horizontal coordinate from exposure, so the association is descriptive rather than an
algebraic restatement of Eq.~\ref{eq:wedge}. Qwen3 is least exposed and OLMoE most exposed; the
causal test of the backbone lever is Section~\ref{sec:resultsII}.}
\label{fig:dilution}
\end{figure}

\paragraph{Backbone share tracks the inversion.} The checkpoints vary from top-$2$-of-$8$ to
top-$8$-of-$128$ routing and include shared and shared-free blocks. Yet Qwen3-30B, the most
fine-grained router, combines one of the largest routing factors ($0.133$ on PubMedQA) with the
smallest exposure ($0.019$). OLMoE shows the reverse ($0.124$ and $0.056$). Directly measuring
$\lVert y\rVert/\lVert\mathrm{res}\rVert$ preserves this ordering (Figure~\ref{fig:dilution}); over
the seven checkpoints its Spearman association with exposure is $0.96$ on PubMedQA and $1.00$ on
GSM8K. The dense-like term $B$ is also larger than $A$ on every checkpoint
($1.6$--$4.1\times$), motivating the output intervention in Section~\ref{sec:resultsII}.

\paragraph{The narrow range has a scale boundary.} We hold Qwen3 expert count and top-$k$ fixed
while moving from 30B to 235B. The routing factor rises to $0.188/0.200$ on PubMedQA/GSM8K, above
the observed range, while exposure remains $0.030$. The same pattern does not repeat in a second
probe: DeepSeek-V2 at 236B remains at or inside the range ($0.114/0.139$) and has exposure
$0.012/0.019$. The two probes establish that the block-level range is not scale-universal, but they
do not isolate a scaling law. Exposure remains within or below the primary observed range in both.

\section{Results II: the causal intervention}
\label{sec:resultsII}

\paragraph{From observation to manipulation.} The wedge is an identity, not a mechanism. We
therefore scale the always-on shared output directly in two confirmatory checkpoints,
DeepSeek-V2-Lite and Gravity. At each block we multiply that output by
$\alpha\in\{1,0.75,0.5,0.25,0\}$, propagate the full forward pass, and remeasure on the fixed base
trajectory. The router is not directly modified; it responds only to the resulting downstream drift.
The $\alpha=1$ no-op reproduces the baseline exactly. Exposure here is measured post-block, and we
interpret the graded response before treating full removal as in-distribution behavior.

\paragraph{Always-on mass contains routing mismatch through coherence.} Removing backbone mass
raises post-block exposure monotonically: from $0.024$ to $0.040$ on DeepSeek-V2-Lite and from
$0.040$ to $0.077$ on Gravity. A frozen-numerator null, which changes only the denominator, is nearly
flat. Instead, $\lVert A\rVert$ rises to $1.57\times$ and $1.69\times$ its baseline as routing
coherence falls. Thus the causal response is not pure denominator dilution. The always-on path
contains mismatch mainly by keeping the two passes' routes coherent.

\begin{figure}[t]
\centering
\includegraphics[width=\linewidth]{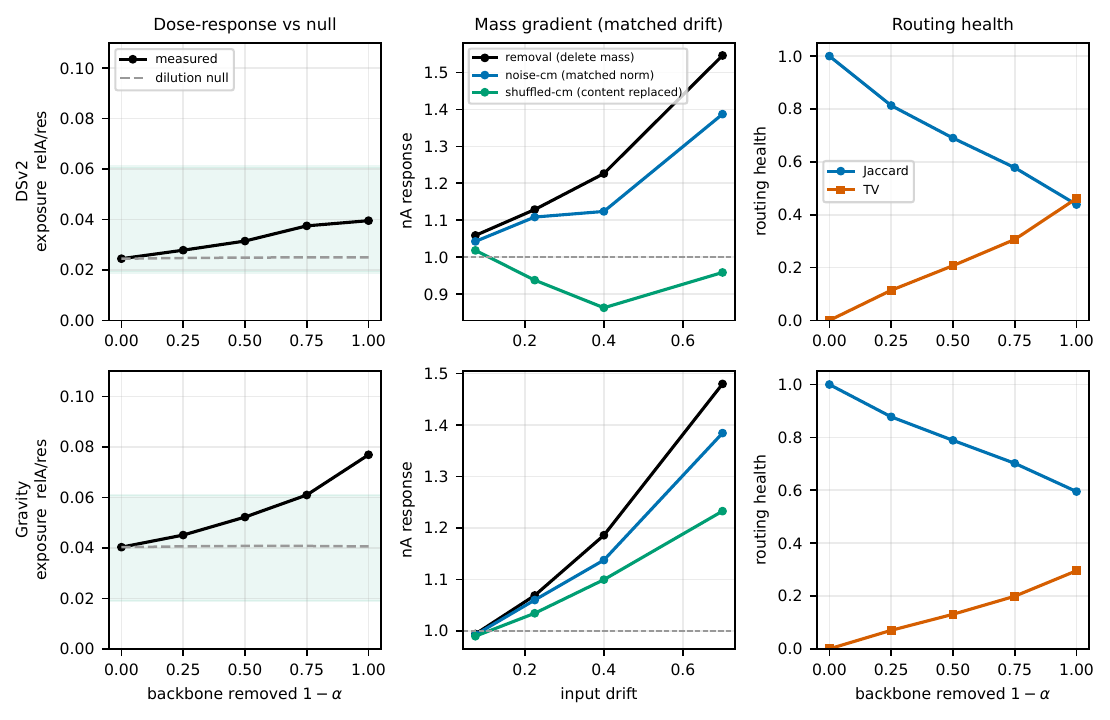}
\caption{Backbone rescaling on the two confirmatory checkpoints: DeepSeek-V2-Lite (top) and Gravity
(bottom). Left: post-block exposure against the frozen-numerator null. Center: routing-coherence
response under the mass-removal, matched-norm, and content-replacement controls at matched input
drift. Right: top-$k$ Jaccard and gate TV across the dose trajectory. Full removal changes routing
substantially; the claim rests on the graded response and control ordering.}
\label{fig:intervention}
\end{figure}

\paragraph{Preserved mass contributes beyond content.} A valid control must change the shared path
in common mode across teacher and student; independent perturbations create an artificial
between-pass mismatch. At matched input drift, replacing shared content while preserving its mass
produces less routing decoherence than a matched-norm off-manifold perturbation, which in turn
produces less than removing the mass (Figure~\ref{fig:intervention}). A descriptive layer-point
bootstrap preserves the three-way ordering in $96\%$ and $98\%$ of resamples; because layers recur
across doses, this is a stability diagnostic rather than a prompt-level interval. The content-preserving arm
is near the no-op on DeepSeek-V2-Lite but reacts on Gravity, so we do not claim complete content
irrelevance. The supported claim is narrower: preserved mass contributes beyond content, and its
dominant measured effect is routing coherence.

\paragraph{Scope and replication.} Full removal substantially changes routing, so the mechanistic
claim rests on the graded trajectory and common-mode ladder, not on treating the knockout as a
natural operating point. Repeating the same $n=100$ protocol on GSM8K reproduces the result on both
checkpoints: exposure rises against a nearly flat null ($72\%$ on DeepSeek-V2-Lite and $106\%$ on
Gravity), and the same control ordering holds. Detailed trajectories and routing-health measurements
are in Appendix~\ref{app:intervention}.

\paragraph{Causal $A/B$ patches connect the instrument to behavior.} In a preregistered test on
OLMoE, DeepSeek-V2-Lite, and Qwen3-30B ($n=200$), we inject the recorded student-anchored term $A$
at every MoE block and compare it with matched-norm random directions and the natural context effect.
Injected $A$ flips $3.1$--$4.0\%$ of greedy tokens, $40$--$45\%$ of the context reference, while
matched-norm noise reproduces $58$--$74\%$ of that flip rate. The routing term therefore has a real
but bounded effect that is largely generic to perturbation magnitude. A GSM8K replication gives the
same qualitative result.

The $B$-side arm closes the attribution: at its own norm, the content term's premium over random
direction is $5.6$--$80\times$, against $1.54$--$2.05\times$ for $A$ on the same checkpoints and
domain. The natural context effect is carried much more strongly by $B$'s direction. Behavioral
effect does not order cleanly by exposure across three checkpoints, so exposure is a screening
coordinate, not a behavioral threshold. Finally, the exact decomposition permits a symmetric
teacher anchor; its norms are similar but its direction was not patched, so the causal claim is
specific to the student-anchored $A$ defined in Eq.~\ref{eq:A}.

\section{Related work}
\label{sec:related}
Sparse MoE layers route each token to a small subset of experts \citep{shazeer2017moe}. This
discrete assignment motivates work on load balance and router stability
\citep{lepikhin2021gshard,fedus2022switch,dai2022stablemoe}. Later studies track router drift during
continual tuning or suppress it through consistency, replay, and compatible expert selection
\citep{xie2026same,hou2026passmoe,guo2026starmoe,liu2026cpmoe,omi2025similarity}. In reinforcement
learning, rollout-to-update routing mismatch can destabilize training
\citep{ma2025r3,dong2026pr2}. Those studies address assignment stability, sample efficiency,
forgetting, or trajectory-level optimization. We address a different estimand: how much a realized
same-weight conditioning mismatch propagates into the residual stream and output. Containment here
does not imply that trajectory-level router instability is harmless.

Shared-expert architectures send every token through always-on computation while routed experts
specialize \citep{dai2024deepseekmoe,deepseekv2}. Sparse upcycling similarly anchors experts to a
dense checkpoint \citep{komatsuzaki2023sparse,he2024upcycling,zhang2024bam}. These designs suggest
that common computation can stabilize behavior, but do not measure the residual exposure of an
actual routing perturbation. Our backbone intervention supplies that measurement and separates a
small denominator-only effect from the larger coherence response.

Measurement work interprets experts, separates routing inputs from ignored content, and evaluates
counterfactual route changes \citep{herbst2026expertstrikes,yang2025moex,ye2026polysemantic,
yoon2026misrouted}. The last of these asks how much an optimized alternative route can move fragile
tokens; our quantity is the realized conditioning-induced route difference. The two are
complementary: a large effect available somewhere in route space does not imply that the route change
produced by a particular training example has that effect.

Same-weight self-distillation uses a model under additional conditioning as its own teacher
\citep{yang2024sdft,shenfeld2026sdft}. That is exactly the two-pass construction measured here.
Distillation from a separate checkpoint \citep{agarwal2024gkd,qwen3_2025,glm45_2025} introduces an
additional weight-induced mismatch, and self-teacher reinforcement learning
\citep{hubotter2026sdpo} introduces a trajectory; both lie outside our single-step scope. We also do
not expect the observed range to transfer unchanged to expert-choice, metadata-routed, or
independently merged experts \citep{zhou2022expertchoice,gururangan2022demix,
li2022branchtrainmerge,wortsman2022modelsoups,ilharco2023taskarithmetic,
sukhbaatar2024branchtrainmix}. The merged-expert probe below tests one such boundary.

\section{Discussion and limitations}
\label{sec:discussion}

\paragraph{From motion to influence.}
Routing motion and routing influence are different quantities. Gate overlap describes whether routes
changed; the decomposition measures the block-output perturbation those changes produced; exposure
measures how much reached the residual stream. The screen links exposure to backbone share, the
rescaling intervention shows that always-on mass also preserves route coherence, and the patches
show that the routing term can change outputs but is smaller and less direction-specific than the
dense-like content term. The result is containment, not harmlessness.

\paragraph{What the observed range means.}
Normalizing by the routed block output reduces sensitivity to gate convention and raw expert-output
scale, but it does not derive the observed width. The two scale probes behave differently, and each
changes several capacity or training axes at once. Together with the merged-expert boundary, they
rule out a universal range without localizing a scaling law. Exposure remains within or below the
primary observed range in both scale probes, but that observation is not a behavioral threshold.

\paragraph{A boundary that fails in the informative direction.}
A merged-expert checkpoint tests a design outside the jointly trained family. Beyonder-4x7B-v3 is a
mergekit-built MoE \citep{goddard2024mergekit,beyonder2024} whose router was not jointly trained with
its four experts. Its routing factor falls to $0.014$ on PubMedQA and $0.011$ on GSM8K, far below
the primary range, while exposure is $0.005$--$0.006$. We predicted that the range would break but
expected an upward break because the design lacks an always-on backbone. The direction was wrong.
Despite ordinary gate movement, reallocating mass changes the block little, consistent with high
similarity among experts derived from the same dense base. Backbone mass can contain a perturbation
after it forms; sufficiently similar expert outputs can suppress it at the source.

\paragraph{Use as a diagnostic, not a threshold.}
A checkpoint-level audit should follow the evidence. First verify that the decomposition reconstructs
the routed-output difference. Then measure exposure rather than inferring influence from router
overlap. If the downstream decision matters, patch the recorded term or run another behavioral
intervention. Exposure within the observed range is evidence of residual containment in this regime,
not proof that behavior is unchanged. The larger and more direction-specific $B$ term also means
that routing-only mitigation should not be assumed to address the dominant context-induced change.

\paragraph{Limits of the evidence.}
The seven open-weight checkpoints form an opportunistic, family-correlated sample rather than a
factorial architecture sweep. Backbone rescaling is confirmatory only for DeepSeek-V2-Lite and
Gravity, whose always-on branches provide a clean scalar handle. The merged boundary is one
checkpoint. The output patches cover three checkpoints, and exposure does not order their
behavioral effects. All-layer patches can also compound recorded local terms differently from the
natural context trajectory, so the matched-norm $A/B$ comparison establishes relative direction
specificity rather than an exact partition of the final output change. The reported behaviors are
next-token quantities, not task-level generation quality. The causal patch uses the student anchor
from Eq.~\ref{eq:A}; the alternate anchor has similar norm but a different vector. The instrument
currently requires reconstruction-valid, full-precision expert execution. Most importantly, the
claims concern one same-weight conditioning mismatch. They do not cover accumulated optimization,
separately parameterized teachers, reinforcement-learning trajectories, or forgetting.

\section{Conclusion}
Same weights do not imply the same computation in an MoE: conditioning can change the route. But
route change is not the same as downstream influence. Our decomposition follows the mismatch from
the gates, through the routed block, into the residual stream. Across the tested checkpoints,
backbone share orders exposure; causal rescaling shows that always-on mass contains mismatch chiefly
by preserving routing coherence; and output patches reveal a bounded, mostly magnitude-generic
routing effect beside a more direction-specific content effect. The narrow block-level range is not
universal; scale and merged-expert designs break it. The durable practical lesson is narrower: use
router divergence to locate change, residual exposure to quantify propagation, and behavioral
intervention to establish consequence.

\section*{Reproducibility statement}
Analysis scripts and stochastic controls use fixed seeds; the frozen analysis-ready captures
support deterministic reduction. Fresh model acquisition is a scientific replication, not a
guaranteed bitwise replay. Code and captures underlying the reported figures and tables are available at
\url{https://github.com/CedricCaruzzo/routing-divergence-moe}. Public model release
names are listed in Appendix~\ref{app:tables}.

\section*{Author Contributions}
Contributor roles are described using the CRediT
taxonomy.\footnote{Contributor Roles Taxonomy (CRediT), an ANSI/NISO standard: \url{https://credit.niso.org}.}

\noindent\textbf{C.C.:} Conceptualization, Methodology, Software, Formal analysis,
Investigation, Validation, Visualization, Writing~--~original draft.\\
\textbf{D.Y.:} Funding acquisition, Resources.\\
\textbf{T.S.K.:} Funding acquisition, Resources.

\bibliographystyle{plainnat}
\bibliography{refs}

\begin{thebibliography}{42}
\providecommand{\natexlab}[1]{#1}
\providecommand{\url}[1]{\texttt{#1}}
\expandafter\ifx\csname urlstyle\endcsname\relax
  \providecommand{\doi}[1]{doi: #1}\else
  \providecommand{\doi}{doi: \begingroup \urlstyle{rm}\Url}\fi

\bibitem[Abdin et~al.(2024)Abdin, Aneja, Awadalla, Awadallah, Awan, Bach,
  Bahree, Bakhtiari, Bao, Behl, et~al.]{abdin2024phi3}
Marah Abdin, Jyoti Aneja, Hany Awadalla, Ahmed Awadallah, Ammar~Ahmad Awan,
  Nguyen Bach, Amit Bahree, Arash Bakhtiari, Jianmin Bao, Harkirat Behl, et~al.
\newblock Phi-3 technical report: A highly capable language model locally on
  your phone.
\newblock \emph{arXiv preprint arXiv:2404.14219}, 2024.
\newblock URL \url{https://arxiv.org/abs/2404.14219}.

\bibitem[Agarwal et~al.(2024)Agarwal, Vieillard, Zhou, Stanczyk, Ramos, Geist,
  and Bachem]{agarwal2024gkd}
Rishabh Agarwal, Nino Vieillard, Yongchao Zhou, Piotr Stanczyk, Sabela Ramos,
  Matthieu Geist, and Olivier Bachem.
\newblock On-policy distillation of language models: Learning from
  self-generated mistakes.
\newblock In \emph{International Conference on Learning Representations
  (ICLR)}, 2024.
\newblock URL \url{https://arxiv.org/abs/2306.13649}.

\bibitem[Austin et~al.(2021)Austin, Odena, Nye, Bosma, Michalewski, Dohan,
  Jiang, Cai, Terry, Le, and Sutton]{austin2021program}
Jacob Austin, Augustus Odena, Maxwell Nye, Maarten Bosma, Henryk Michalewski,
  David Dohan, Ellen Jiang, Carrie Cai, Michael Terry, Quoc~V. Le, and Charles
  Sutton.
\newblock Program synthesis with large language models.
\newblock \emph{arXiv preprint arXiv:2108.07732}, 2021.
\newblock URL \url{https://arxiv.org/abs/2108.07732}.

\bibitem[Cobbe et~al.(2021)Cobbe, Kosaraju, Bavarian, Chen, Jun, Kaiser,
  Plappert, Tworek, Hilton, Nakano, Hesse, and Schulman]{cobbe2021gsm8k}
Karl Cobbe, Vineet Kosaraju, Mohammad Bavarian, Mark Chen, Heewoo Jun, Lukasz
  Kaiser, Matthias Plappert, Jerry Tworek, Jacob Hilton, Reiichiro Nakano,
  Christopher Hesse, and John Schulman.
\newblock Training verifiers to solve math word problems.
\newblock \emph{arXiv preprint arXiv:2110.14168}, 2021.
\newblock URL \url{https://arxiv.org/abs/2110.14168}.

\bibitem[Dai et~al.(2022)Dai, Dong, Ma, Zheng, Sui, Chang, and
  Wei]{dai2022stablemoe}
Damai Dai, Li~Dong, Shuming Ma, Bo~Zheng, Zhifang Sui, Baobao Chang, and Furu
  Wei.
\newblock Stablemoe: Stable routing strategy for mixture of experts.
\newblock In \emph{Proceedings of the 60th Annual Meeting of the Association
  for Computational Linguistics (ACL)}, 2022.
\newblock URL \url{https://arxiv.org/abs/2204.08396}.

\bibitem[Dai et~al.(2024)Dai, Deng, Zhao, Xu, Gao, Chen, Li, Zeng, Yu,
  et~al.]{dai2024deepseekmoe}
Damai Dai, Chengqi Deng, Chenggang Zhao, R.X. Xu, Huazuo Gao, Deli Chen, Jiashi
  Li, Wangding Zeng, Xingkai Yu, et~al.
\newblock {DeepSeekMoE}: Towards ultimate expert specialization in
  mixture-of-experts language models.
\newblock In \emph{Proceedings of the 62nd Annual Meeting of the Association
  for Computational Linguistics (ACL)}, 2024.
\newblock URL \url{https://arxiv.org/abs/2401.06066}.

\bibitem[{DeepSeek-AI}(2024)]{deepseekv2}
{DeepSeek-AI}.
\newblock Deepseek-v2: A strong, economical, and efficient mixture-of-experts
  language model.
\newblock \emph{arXiv preprint arXiv:2405.04434}, 2024.
\newblock URL \url{https://arxiv.org/abs/2405.04434}.

\bibitem[Dong et~al.(2026)Dong, Chen, Jia, Liu, Wu, Di, Wu, Liu, Liu, Barsoum,
  Metaxas, and Wang]{dong2026pr2}
Daize Dong, Junlin Chen, Haolong Jia, Jiang Liu, Jiawei Wu, Huanwei Di, Jialian
  Wu, Zhengzhong Liu, Zicheng Liu, Emad Barsoum, Dimitris~N. Metaxas, and
  Hongyi Wang.
\newblock {PR2}: Predictive routing replay for moe-based llm reinforcement
  learning.
\newblock \emph{arXiv preprint arXiv:2606.00395}, 2026.
\newblock URL \url{https://arxiv.org/abs/2606.00395}.

\bibitem[Fedus et~al.(2022)Fedus, Zoph, and Shazeer]{fedus2022switch}
William Fedus, Barret Zoph, and Noam Shazeer.
\newblock Switch transformers: Scaling to trillion parameter models with simple
  and efficient sparsity.
\newblock \emph{Journal of Machine Learning Research}, 23\penalty0
  (120):\penalty0 1--39, 2022.
\newblock URL \url{https://arxiv.org/abs/2101.03961}.

\bibitem[{GLM Team}(2025)]{glm45_2025}
{GLM Team}.
\newblock Glm-4.5: Agentic, reasoning, and coding (arc) foundation models.
\newblock \emph{arXiv preprint arXiv:2508.06471}, 2025.
\newblock URL \url{https://arxiv.org/abs/2508.06471}.

\bibitem[Goddard et~al.(2024)Goddard, Siriwardhana, Ehghaghi, Meyers,
  Karpukhin, Benedict, McQuade, and Solawetz]{goddard2024mergekit}
Charles Goddard, Shamane Siriwardhana, Malikeh Ehghaghi, Luke Meyers, Vladimir
  Karpukhin, Brian Benedict, Mark McQuade, and Jacob Solawetz.
\newblock Arcee's {MergeKit}: A toolkit for merging large language models.
\newblock \emph{arXiv preprint arXiv:2403.13257}, 2024.
\newblock URL \url{https://arxiv.org/abs/2403.13257}.

\bibitem[Guo et~al.(2026)Guo, Cheng, Zhou, and Zhang]{guo2026starmoe}
Zirui Guo, Quan Cheng, Da-Wei Zhou, and Lijun Zhang.
\newblock Stable routing for mixture-of-experts in class-incremental learning.
\newblock \emph{arXiv preprint arXiv:2605.17571}, 2026.
\newblock URL \url{https://arxiv.org/abs/2605.17571}.

\bibitem[Gururangan et~al.(2022)Gururangan, Lewis, Holtzman, Smith, and
  Zettlemoyer]{gururangan2022demix}
Suchin Gururangan, Mike Lewis, Ari Holtzman, Noah~A. Smith, and Luke
  Zettlemoyer.
\newblock Demix layers: Disentangling domains for modular language modeling.
\newblock In \emph{Proceedings of the 2022 Conference of the North American
  Chapter of the Association for Computational Linguistics (NAACL)}, pages
  5557--5576, 2022.
\newblock URL \url{https://aclanthology.org/2022.naacl-main.407/}.

\bibitem[He et~al.(2024)He, Khattar, Prenger, Korthikanti, Yan, Liu, Fan,
  Aithal, Shoeybi, and Catanzaro]{he2024upcycling}
Ethan He, Abhinav Khattar, Ryan Prenger, Vijay Korthikanti, Zijie Yan, Tong
  Liu, Shiqing Fan, Ashwath Aithal, Mohammad Shoeybi, and Bryan Catanzaro.
\newblock Upcycling large language models into mixture of experts.
\newblock \emph{arXiv preprint arXiv:2410.07524}, 2024.
\newblock URL \url{https://arxiv.org/abs/2410.07524}.

\bibitem[Herbst et~al.(2026)Herbst, Wermter, and Lee]{herbst2026expertstrikes}
Jeremy Herbst, Stefan Wermter, and Jae~Hee Lee.
\newblock The expert strikes back: Interpreting mixture-of-experts language
  models at expert level.
\newblock In \emph{Proceedings of the 43rd International Conference on Machine
  Learning (ICML)}, 2026.
\newblock URL \url{https://arxiv.org/abs/2604.02178}.

\bibitem[Hou et~al.(2026)Hou, Guo, Ma, Sun, Yang, and Wang]{hou2026passmoe}
Zhiyan Hou, Haiyun Guo, Haokai Ma, Yandu Sun, Yonghui Yang, and Jinqiao Wang.
\newblock Pass-moe: Mitigating misaligned co-drift among router and experts via
  pathway activation subspaces for continual learning.
\newblock \emph{arXiv preprint arXiv:2601.13020}, 2026.
\newblock URL \url{https://arxiv.org/abs/2601.13020}.

\bibitem[H{\"u}botter et~al.(2026)H{\"u}botter, L{\"u}beck, Behric, Baumann,
  Bagatella, Marta, Hakimi, Shenfeld, Kleine~Buening, Guestrin, and
  Krause]{hubotter2026sdpo}
Jonas H{\"u}botter, Frederike L{\"u}beck, Lejs Behric, Anton Baumann, Marco
  Bagatella, Daniel Marta, Ido Hakimi, Idan Shenfeld, Thomas Kleine~Buening,
  Carlos Guestrin, and Andreas Krause.
\newblock Reinforcement learning via self-distillation.
\newblock \emph{arXiv preprint arXiv:2601.20802}, 2026.
\newblock URL \url{https://arxiv.org/abs/2601.20802}.

\bibitem[Ilharco et~al.(2023)Ilharco, Ribeiro, Wortsman, Gururangan, Schmidt,
  Hajishirzi, and Farhadi]{ilharco2023taskarithmetic}
Gabriel Ilharco, Marco~Tulio Ribeiro, Mitchell Wortsman, Suchin Gururangan,
  Ludwig Schmidt, Hannaneh Hajishirzi, and Ali Farhadi.
\newblock Editing models with task arithmetic.
\newblock In \emph{International Conference on Learning Representations
  (ICLR)}, 2023.
\newblock URL \url{https://arxiv.org/abs/2212.04089}.

\bibitem[Jiang et~al.(2024)Jiang, Sablayrolles, Roux, Mensch, Savary, Bamford,
  Chaplot, Casas, Hanna, Bressand, et~al.]{jiang2024mixtral}
Albert~Q. Jiang, Alexandre Sablayrolles, Antoine Roux, Arthur Mensch, Blanche
  Savary, Chris Bamford, Devendra~Singh Chaplot, Diego de~las Casas, Emma~Bou
  Hanna, Florian Bressand, et~al.
\newblock Mixtral of experts.
\newblock \emph{arXiv preprint arXiv:2401.04088}, 2024.
\newblock URL \url{https://arxiv.org/abs/2401.04088}.

\bibitem[Jin et~al.(2019)Jin, Dhingra, Liu, Cohen, and Lu]{jin2019pubmedqa}
Qiao Jin, Bhuwan Dhingra, Zhengping Liu, William~W. Cohen, and Xinghua Lu.
\newblock {PubMedQA}: A dataset for biomedical research question answering.
\newblock In \emph{Proceedings of the 2019 Conference on Empirical Methods in
  Natural Language Processing and the 9th International Joint Conference on
  Natural Language Processing (EMNLP-IJCNLP)}, pages 2567--2577, 2019.
\newblock \doi{10.18653/v1/D19-1259}.
\newblock URL \url{https://aclanthology.org/D19-1259/}.

\bibitem[Komatsuzaki et~al.(2023)Komatsuzaki, Puigcerver, Lee-Thorp,
  Riquelme~Ruiz, Mustafa, Ainslie, Tay, Dehghani, and
  Houlsby]{komatsuzaki2023sparse}
Aran Komatsuzaki, Joan Puigcerver, James Lee-Thorp, Carlos Riquelme~Ruiz, Basil
  Mustafa, Joshua Ainslie, Yi~Tay, Mostafa Dehghani, and Neil Houlsby.
\newblock Sparse upcycling: Training mixture-of-experts from dense checkpoints.
\newblock In \emph{International Conference on Learning Representations
  (ICLR)}, 2023.
\newblock URL \url{https://arxiv.org/abs/2212.05055}.

\bibitem[Labonne(2024)]{beyonder2024}
Maxime Labonne.
\newblock Beyonder-4x7b-v3.
\newblock Hugging Face model card, 2024.
\newblock URL \url{https://huggingface.co/mlabonne/Beyonder-4x7B-v3}.

\bibitem[Lepikhin et~al.(2021)Lepikhin, Lee, Xu, Chen, Firat, Huang, Krikun,
  Shazeer, and Chen]{lepikhin2021gshard}
Dmitry Lepikhin, HyoukJoong Lee, Yuanzhong Xu, Dehao Chen, Orhan Firat, Yanping
  Huang, Maxim Krikun, Noam Shazeer, and Zhifeng Chen.
\newblock Gshard: Scaling giant models with conditional computation and
  automatic sharding.
\newblock In \emph{International Conference on Learning Representations
  (ICLR)}, 2021.
\newblock URL \url{https://arxiv.org/abs/2006.16668}.

\bibitem[Li et~al.(2022)Li, Gururangan, Dettmers, Lewis, Althoff, Smith, and
  Zettlemoyer]{li2022branchtrainmerge}
Margaret Li, Suchin Gururangan, Tim Dettmers, Mike Lewis, Tim Althoff, Noah~A.
  Smith, and Luke Zettlemoyer.
\newblock Branch-train-merge: Embarrassingly parallel training of expert
  language models.
\newblock \emph{arXiv preprint arXiv:2208.03306}, 2022.
\newblock URL \url{https://arxiv.org/abs/2208.03306}.

\bibitem[Liu et~al.(2026)Liu, Nguyen, and Salim]{liu2026cpmoe}
Yang Liu, Toan Nguyen, and Flora~D. Salim.
\newblock Cp-moe: Consistency-preserving mixture-of-experts for continual
  learning.
\newblock \emph{arXiv preprint arXiv:2605.20247}, 2026.
\newblock URL \url{https://arxiv.org/abs/2605.20247}.

\bibitem[Ma et~al.(2025)Ma, Zhang, Zhao, Song, Wang, Sui, and Luo]{ma2025r3}
Wenhan Ma, Hailin Zhang, Liang Zhao, Yifan Song, Yudong Wang, Zhifang Sui, and
  Fuli Luo.
\newblock Stabilizing moe reinforcement learning by aligning training and
  inference routers.
\newblock \emph{arXiv preprint arXiv:2510.11370}, 2025.
\newblock URL \url{https://arxiv.org/abs/2510.11370}.

\bibitem[Muennighoff et~al.(2024)Muennighoff, Soldaini, Groeneveld, Lo,
  Morrison, Min, Shi, Walsh, Tafjord, Lambert, et~al.]{muennighoff2024olmoe}
Niklas Muennighoff, Luca Soldaini, Dirk Groeneveld, Kyle Lo, Jacob Morrison,
  Sewon Min, Weijia Shi, Pete Walsh, Oyvind Tafjord, Nathan Lambert, et~al.
\newblock Olmoe: Open mixture-of-experts language models.
\newblock \emph{arXiv preprint arXiv:2409.02060}, 2024.
\newblock URL \url{https://arxiv.org/abs/2409.02060}.

\bibitem[Omi et~al.(2025)Omi, Sen, and Farhadi]{omi2025similarity}
Nabil Omi, Siddhartha Sen, and Ali Farhadi.
\newblock Load balancing mixture of experts with similarity preserving routers.
\newblock \emph{arXiv preprint arXiv:2506.14038}, 2025.
\newblock URL \url{https://arxiv.org/abs/2506.14038}.

\bibitem[{Qwen Team}(2024)]{qwen2024qwen15moe}
{Qwen Team}.
\newblock Qwen1.5-moe: Matching 7b model performance with 1/3 activated
  parameters.
\newblock Qwen blog, 2024.
\newblock URL \url{https://qwenlm.github.io/blog/qwen-moe/}.
\newblock Accessed 2026-07-22.

\bibitem[{Qwen Team}(2025)]{qwen3_2025}
{Qwen Team}.
\newblock Qwen3 technical report.
\newblock \emph{arXiv preprint arXiv:2505.09388}, 2025.
\newblock URL \url{https://arxiv.org/abs/2505.09388}.

\bibitem[Shazeer et~al.(2017)Shazeer, Mirhoseini, Maziarz, Davis, Le, Hinton,
  and Dean]{shazeer2017moe}
Noam Shazeer, Azalia Mirhoseini, Krzysztof Maziarz, Andy Davis, Quoc Le,
  Geoffrey Hinton, and Jeff Dean.
\newblock Outrageously large neural networks: The sparsely-gated
  mixture-of-experts layer.
\newblock In \emph{International Conference on Learning Representations
  (ICLR)}, 2017.
\newblock URL \url{https://arxiv.org/abs/1701.06538}.

\bibitem[Shenfeld et~al.(2026)Shenfeld, Damani, H{\"u}botter, and
  Agrawal]{shenfeld2026sdft}
Idan Shenfeld, Mehul Damani, Jonas H{\"u}botter, and Pulkit Agrawal.
\newblock Self-distillation enables continual learning.
\newblock \emph{arXiv preprint arXiv:2601.19897}, 2026.
\newblock URL \url{https://arxiv.org/abs/2601.19897}.

\bibitem[Sukhbaatar et~al.(2024)Sukhbaatar, Golovneva, Sharma, Xu, Lin,
  Rozi{\`e}re, Kahn, Li, Yih, Weston, and Li]{sukhbaatar2024branchtrainmix}
Sainbayar Sukhbaatar, Olga Golovneva, Vasu Sharma, Hu~Xu, Xi~Victoria Lin,
  Baptiste Rozi{\`e}re, Jacob Kahn, Daniel Li, Wen-tau Yih, Jason Weston, and
  Xian Li.
\newblock Branch-train-mix: Mixing expert llms into a mixture-of-experts llm.
\newblock \emph{arXiv preprint arXiv:2403.07816}, 2024.
\newblock URL \url{https://arxiv.org/abs/2403.07816}.

\bibitem[{Trillion Labs}(2026)]{trillion2026gravity}
{Trillion Labs}.
\newblock Gravity-16b-a3b-preview.
\newblock Hugging Face model card, 2026.
\newblock URL
  \url{https://huggingface.co/trillionlabs/Gravity-16B-A3B-Preview}.

\bibitem[Wortsman et~al.(2022)Wortsman, Ilharco, Gadre, Roelofs, Gontijo-Lopes,
  Morcos, Namkoong, Farhadi, Carmon, Kornblith, and
  Schmidt]{wortsman2022modelsoups}
Mitchell Wortsman, Gabriel Ilharco, Samir~Yitzhak Gadre, Rebecca Roelofs,
  Raphael Gontijo-Lopes, Ari~S. Morcos, Hongseok Namkoong, Ali Farhadi, Yair
  Carmon, Simon Kornblith, and Ludwig Schmidt.
\newblock Model soups: Averaging weights of multiple fine-tuned models improves
  accuracy without increasing inference time.
\newblock In \emph{Proceedings of the 39th International Conference on Machine
  Learning (ICML)}, 2022.
\newblock URL \url{https://proceedings.mlr.press/v162/wortsman22a.html}.

\bibitem[Xie et~al.(2026)Xie, Tang, Shi, Ye, Zhan, and Zhou]{xie2026same}
Zhen-Hao Xie, Jun-Tao Tang, Yu-Cheng Shi, Han-Jia Ye, De-Chuan Zhan, and Da-Wei
  Zhou.
\newblock Same: Stabilized mixture-of-experts for multimodal continual
  instruction tuning.
\newblock \emph{arXiv preprint arXiv:2602.01990}, 2026.
\newblock URL \url{https://arxiv.org/abs/2602.01990}.

\bibitem[Yang et~al.(2025)Yang, Venhoff, Khakzar, Schroeder~de Witt, Dokania,
  Bibi, and Torr]{yang2025moex}
Xingyi Yang, Constantin Venhoff, Ashkan Khakzar, Christian Schroeder~de Witt,
  Puneet~K. Dokania, Adel Bibi, and Philip Torr.
\newblock Mixture of experts made intrinsically interpretable.
\newblock In \emph{Proceedings of the 42nd International Conference on Machine
  Learning (ICML)}, 2025.
\newblock URL \url{https://arxiv.org/abs/2503.07639}.

\bibitem[Yang et~al.(2024)Yang, Liu, Pang, Wang, Feng, Zhu, and
  Chen]{yang2024sdft}
Zhaorui Yang, Qian Liu, Tianyu Pang, Han Wang, Haozhe Feng, Minfeng Zhu, and
  Wei Chen.
\newblock Self-distillation bridges distribution gap in language model
  fine-tuning.
\newblock In \emph{Proceedings of the 62nd Annual Meeting of the Association
  for Computational Linguistics (ACL)}, 2024.
\newblock URL \url{https://arxiv.org/abs/2402.13669}.

\bibitem[Ye et~al.(2026)Ye, Yuan, and Sharkey]{ye2026polysemantic}
Charles Ye, Bo~Yuan, and Lee Sharkey.
\newblock Polysemantic experts, monosemantic paths: Routing as control in moes.
\newblock \emph{arXiv preprint arXiv:2604.17837}, 2026.
\newblock URL \url{https://arxiv.org/abs/2604.17837}.

\bibitem[Yoon et~al.(2026)Yoon, Wang, Chen, and Ok]{yoon2026misrouted}
Youngsik Yoon, Siwei Wang, Wei Chen, and Jungseul Ok.
\newblock When are experts misrouted? counterfactual routing analysis in
  mixture-of-experts language models.
\newblock \emph{arXiv preprint arXiv:2605.07260}, 2026.
\newblock URL \url{https://arxiv.org/abs/2605.07260}.

\bibitem[Zhang et~al.(2024)Zhang, Gritsch, Gnaneshwar, Guo, Cairuz, Venkitesh,
  Foerster, Blunsom, Ruder, Ustun, and Locatelli]{zhang2024bam}
Qizhen Zhang, Nikolas Gritsch, Dwaraknath Gnaneshwar, Simon Guo, David Cairuz,
  Bharat Venkitesh, Jakob Foerster, Phil Blunsom, Sebastian Ruder, Ahmet Ustun,
  and Acyr Locatelli.
\newblock Bam! just like that: Simple and efficient parameter upcycling for
  mixture of experts.
\newblock \emph{arXiv preprint arXiv:2408.08274}, 2024.
\newblock URL \url{https://arxiv.org/abs/2408.08274}.

\bibitem[Zhou et~al.(2022)Zhou, Lei, Liu, Du, Huang, Zhao, Dai, Chen, Le, and
  Laudon]{zhou2022expertchoice}
Yanqi Zhou, Tao Lei, Hanxiao Liu, Nan Du, Yanping Huang, Vincent Zhao,
  Andrew~M. Dai, Zhifeng Chen, Quoc~V. Le, and James Laudon.
\newblock Mixture-of-experts with expert choice routing.
\newblock In \emph{Advances in Neural Information Processing Systems
  (NeurIPS)}, 2022.
\newblock URL \url{https://arxiv.org/abs/2202.09368}.

\end{thebibliography}

\appendix

\section{Derivation and anchor choice}\label{app:derivation}
Starting from $\Delta_{\mathrm{routed}} = \sum_j g^T_j f_j(h^T) - \sum_j g^S_j f_j(h^S)$, add and
subtract $\sum_j g^T_j f_j(h^S)$:
\begin{equation*}
  \Delta_{\mathrm{routed}}
   = \underbrace{\textstyle\sum_j (g^T_j - g^S_j) f_j(h^S)}_{A}
   + \underbrace{\textstyle\sum_j g^T_j \big(f_j(h^T) - f_j(h^S)\big)}_{B}.
\end{equation*}
The split is not unique: adding and subtracting $\sum_j g^S_j f_j(h^T)$ instead gives $A' = \sum_j
(g^T_j - g^S_j) f_j(h^T)$ and $B' = \sum_j g^S_j (f_j(h^T) - f_j(h^S))$, evaluating the gate
difference against the teacher hidden state. We use the $h^S$ anchor throughout (the routing
difference applied to the state the student produced). The two anchors agree exactly when
$f_j(h^T) = f_j(h^S)$; their difference, $A - A' = -\sum_j (g^T_j - g^S_j)\big(f_j(h^T) -
f_j(h^S)\big)$, is the gate-difference-weighted content shift. An instrumented re-run of the
PubMedQA screen quantifies it: the alternate-anchor magnitude agrees with the reported one to
within a few percent per layer (median per-layer $\lVert A'\rVert/\lVert A\rVert$ of $0.96$ to
$0.99$ across the seven models), and the routing factor recomputed under $A'$ moves by at most
$4.6\%$ per model (range $0.084$--$0.127$ against $0.085$--$0.133$, same ordering), so the
norm-based quantities we report are anchor-robust. The difference \emph{tensor} is not small
(median per-layer $\lVert A - A'\rVert/\lVert A\rVert$ of $0.36$ to $0.65$): the anchors agree in
magnitude, not as vectors, so a directional analysis of $A$ would need to fix the convention. A
second-domain check (DeepSeek-V2-Lite, GSM8K) reproduces the picture (medians $1.00$ and $0.41$),
and a held-out scale probe remains within the same norm-sensitivity range.
(Both splits are exact identities, so reconstruction closure does not distinguish them.) All
reported quantities use the $h^S$ anchor consistently.
$A$ uses the raw realized gates $g$, so the reconstruction is exact only when the applied gate
convention is reproduced. Quantized expert execution that fails this gate is excluded.

\section{Uncertainty and boundary checkpoints}\label{app:tables}
The public release names for the primary screen are Qwen3-30B-A3B,
Mixtral-8x7B-Instruct-v0.1, OLMoE-1B-7B-0924-Instruct, DeepSeek-V2-Lite-Chat,
Gravity-16B-A3B-Preview, Qwen1.5-MoE-A2.7B-Chat, and Phi-3.5-MoE-instruct. Boundary probes use
Qwen3-235B-A22B, DeepSeek-V2-Chat, and Beyonder-4x7B-v3.

Table~\ref{tab:appci} gives the prompt-level $95\%$ bootstrap CIs behind the Figure~\ref{fig:invariance}
whiskers and the Table~\ref{tab:master} footnote. We resample the $n{=}200$ probe prompts ($2000$
resamples), recompute the per-layer-first ratio on each resample, and take the $2.5$ and $97.5$
percentiles; the full-data reduction reproduces the committed headline exactly. The per-model
intervals pin each checkpoint tightly relative to the observed two-domain range. They quantify
sampling uncertainty on each checkpoint, not uncertainty over a population of architectures.

\begin{table}[t]
\centering
\caption{Prompt-level $95\%$ bootstrap CIs (resampling the $n{=}200$ probe prompts, $2000$ resamples)
for the two headline ratios, both domains. These intervals quantify prompt-sampling uncertainty for
each checkpoint rather than variation across checkpoints.}
\label{tab:appci}
\scriptsize
\setlength{\tabcolsep}{3.5pt}
\begin{tabular}{lcccc}
\toprule
 & \multicolumn{2}{c}{PubMedQA} & \multicolumn{2}{c}{GSM8K} \\
\cmidrule(lr){2-3}\cmidrule(lr){4-5}
Model & \relAblock{} CI & \relAres{} CI & \relAblock{} CI & \relAres{} CI \\
\midrule
Qwen3-30B-A3B    & [0.131, 0.135] & [0.019, 0.020] & [0.137, 0.141] & [0.021, 0.021] \\
Mixtral-8x7B     & [0.091, 0.096] & [0.021, 0.023] & [0.108, 0.111] & [0.028, 0.029] \\
OLMoE-1B-7B      & [0.121, 0.127] & [0.055, 0.058] & [0.120, 0.125] & [0.060, 0.062] \\
DeepSeek-V2-Lite & [0.088, 0.092] & [0.026, 0.027] & [0.100, 0.104] & [0.032, 0.033] \\
Gravity-16B-A3B & [0.097, 0.099] & [0.044, 0.045] & [0.107, 0.109] & [0.052, 0.053] \\
Qwen1.5-MoE-A2.7B & [0.084, 0.087] & [0.035, 0.037] & [0.100, 0.104] & [0.047, 0.049] \\
Phi-3.5-MoE      & [0.104, 0.109] & [0.022, 0.024] & [0.108, 0.112] & [0.026, 0.027] \\
\midrule
\multicolumn{5}{l}{\emph{Scale probe (outside the primary range):}} \\
Qwen3-235B-A22B  & [0.186, 0.191] & [0.029, 0.030] & [0.197, 0.203] & [0.030, 0.031] \\
\bottomrule
\end{tabular}
\end{table}

\begin{table}[t]
\centering
\caption{Boundary checkpoints, shown as PubMedQA/GSM8K. Family pairs are descriptive scale probes;
the merged checkpoint tests transfer outside jointly trained expert families. Exposure uses the
pre-block screen convention.}
\label{tab:appboundaries}
\small
\begin{tabular}{lcc}
\toprule
Checkpoint & \relAblock & \relAres \\
\midrule
Qwen3-30B-A3B      & 0.133/0.139 & 0.019/0.021 \\
Qwen3-235B-A22B    & 0.188/0.200 & 0.030/0.030 \\
DeepSeek-V2-Lite   & 0.090/0.102 & 0.027/0.033 \\
DeepSeek-V2 (236B) & 0.114/0.139 & 0.012/0.019 \\
Beyonder-4x7B-v3   & 0.014/0.011 & 0.006/0.005 \\
\bottomrule
\end{tabular}
\end{table}

\section{Robustness and exclusions}\label{app:pathologies}
\paragraph{Aggregation order.} Averaging tokenwise ratios overweights tokens with small residual
norms. Relative to the per-layer-first estimator, it inflates exposure by $20$--$38\%$ on OLMoE and
DeepSeek-V2-Lite across both domains. It also widens the seven-checkpoint ranges to $0.089$--$0.150$
for the routing factor and $0.020$--$0.073$ for exposure, without changing the central ordering. We
therefore use the preregistered per-layer-first reduction throughout.

\paragraph{Reconstruction validity.} All seven primary checkpoints pass the identity-closure gate.
A quantized checkpoint produced reconstruction error $1.06$ and is excluded from every result that
uses $A$ or $B$. This exclusion defines the operating envelope of the instrument, not additional
evidence for the containment claim.

\paragraph{Prompt format and held-out domain.} Removing the chat template moved the routing factor by
$0.007$ on Mixtral and $0.017$ on Qwen3-30B, both with valid reconstruction. This bounds format
sensitivity on two checkpoints but does not prove format invariance. A held-out MBPP code screen
\citep{austin2021program} did not
define the primary range: six of seven checkpoints remained inside it, while OLMoE was marginally
below ($0.082$ against the $0.085$ floor) with reconstruction error $0.020$ at the gate. We draw no
further inference from that third-domain screen.

\section{Intervention details and stopped extension}\label{app:intervention}
Tables~\ref{tab:appdose} and~\ref{tab:appladder} report the PubMedQA dose trajectory and matched-drift
controls for the two confirmatory checkpoints. Exposure is post-block, matching the intervention
target; the screen's pre-block convention is not mixed into these tables.

\begin{table}[t]
\centering
\caption{PubMedQA backbone-rescaling dose-response. The null freezes the routing numerator; nA is
$\lVert A(\alpha)\rVert/\lVert A(1)\rVert$; Jaccard is routing overlap with $\alpha=1$.
DeepSeek-V2-Lite uses $n=200$ probe pairs and Gravity $n=100$.}
\label{tab:appdose}
\small
\begin{tabular}{llcccc}
\toprule
Model & $\alpha$ & \relAres{} post & null & nA & Jaccard \\
\midrule
\multirow{5}{*}{DeepSeek-V2-Lite}
 & 1.00 & 0.0245 & 0.0245 & 1.000 & 1.000 \\
 & 0.75 & 0.0278 & 0.0247 & 1.095 & 0.813 \\
 & 0.50 & 0.0315 & 0.0249 & 1.209 & 0.690 \\
 & 0.25 & 0.0375 & 0.0250 & 1.407 & 0.578 \\
 & 0.00 & 0.0395 & 0.0250 & 1.567 & 0.438 \\
\midrule
\multirow{5}{*}{Gravity}
 & 1.00 & 0.0403 & 0.0403 & 1.000 & 1.000 \\
 & 0.75 & 0.0451 & 0.0406 & 1.068 & 0.878 \\
 & 0.50 & 0.0522 & 0.0408 & 1.174 & 0.789 \\
 & 0.25 & 0.0610 & 0.0408 & 1.320 & 0.702 \\
 & 0.00 & 0.0769 & 0.0406 & 1.690 & 0.595 \\
\bottomrule
\end{tabular}
\end{table}

\begin{table}[t]
\centering
\caption{PubMedQA matched-drift control ladder. Values are mean routing-coherence response nA by
input-drift bin, pooled over intervening doses. At high drift, preserving shared-path mass while
replacing content produces a smaller response than matched-norm noise or mass removal.}
\label{tab:appladder}
\small
\begin{tabular}{llcccc}
\toprule
Model & arm & [0.00,0.15) & [0.15,0.30) & [0.30,0.50) & [0.50,0.90) \\
\midrule
\multirow{3}{*}{DeepSeek-V2-Lite}
 & content replaced & 1.018 & 0.938 & 0.863 & 0.958 \\
 & matched noise    & 1.043 & 1.109 & 1.124 & 1.387 \\
 & mass removed     & 1.059 & 1.129 & 1.226 & 1.546 \\
\midrule
\multirow{3}{*}{Gravity}
 & content replaced & 0.990 & 1.034 & 1.099 & 1.233 \\
 & matched noise    & 0.992 & 1.060 & 1.138 & 1.384 \\
 & mass removed     & 0.994 & 1.069 & 1.186 & 1.480 \\
\bottomrule
\end{tabular}
\end{table}

\paragraph{Second domain.} The same $n=100$ protocol on GSM8K reproduces the monotone response
against a nearly flat null and the high-drift control ordering on both confirmatory checkpoints.

\begin{table}[t]
\centering
\caption{GSM8K dose-response and the matched-drift ladder's highest bin. The ladder entries are
content replacement / matched noise / mass removal.}
\label{tab:appgsm}
\small
\begin{tabular}{llccccc}
\toprule
Model & $\alpha$ & \relAres{} post & null & nA & Jaccard & ladder [0.50,0.90) \\
\midrule
\multirow{5}{*}{DeepSeek-V2-Lite}
 & 1.00 & 0.0303 & 0.0303 & 1.000 & 1.000 & \\
 & 0.75 & 0.0364 & 0.0307 & 1.153 & 0.803 & \\
 & 0.50 & 0.0412 & 0.0309 & 1.307 & 0.678 & 1.11 / 1.52 / 1.62 \\
 & 0.25 & 0.0502 & 0.0310 & 1.532 & 0.566 & \\
 & 0.00 & 0.0521 & 0.0310 & 1.715 & 0.420 & \\
\midrule
\multirow{5}{*}{Gravity}
 & 1.00 & 0.0495 & 0.0495 & 1.000 & 1.000 & \\
 & 0.75 & 0.0571 & 0.0501 & 1.089 & 0.855 & \\
 & 0.50 & 0.0668 & 0.0505 & 1.207 & 0.755 & 1.24 / 1.39 / 1.44 \\
 & 0.25 & 0.0805 & 0.0507 & 1.408 & 0.657 & \\
 & 0.00 & 0.1020 & 0.0506 & 1.874 & 0.537 & \\
\bottomrule
\end{tabular}
\end{table}

\paragraph{Stopped descriptive extension.} Moonlight-16B-A3B was preregistered as a third
always-on-branch intervention with an all-dose monotonicity gate and a parking rule. Its screened
routing factor was below the primary range on both domains. Exposure then turned over at full
removal on both domains, so the stop gate fired. Accordingly, Moonlight is excluded from the
confirmatory mechanism claim. We report the stopped trajectory only to document that decision; no
conclusion rests on its healthy-dose behavior.

\begin{table}[t]
\centering
\caption{Moonlight stopped extension. The screen factor is pre-block ($n=200$); the dose trajectory
is post-block exposure ($n=100$).}
\label{tab:appmoonlight}
\small
\begin{tabular}{lcccccc}
\toprule
Domain & screen factor & $\alpha=1$ & 0.75 & 0.50 & 0.25 & 0 \\
\midrule
PubMedQA & 0.052 & 0.0241 & 0.0297 & 0.0443 & 0.0819 & 0.0520 \\
GSM8K    & 0.059 & 0.0280 & 0.0356 & 0.0583 & 0.1156 & 0.0850 \\
\bottomrule
\end{tabular}
\end{table}

\FloatBarrier
\section{Wedge consistency}\label{app:wedge}
The wedge identity \eqref{eq:wedge} is exact per layer per token by construction. As an internal
check we recompute the per-layer-first \relAres{} as the mean over layers of the per-layer ratio and
compare it to the stored headline value; the two agree to within $10^{-16}$ across all models and
both domains. Figure~\ref{fig:dilution} instead plots backbone share computed directly from the same
stored sums as
$\operatorname{mean}_{\ell}(\sum\lVert y\rVert/\sum\lVert\mathrm{res}\rVert)$. Because this is a
different aggregate reduction, its product with the plotted routing factor need not reproduce the
aggregate exposure exactly. Identity-implied shares are $3$--$17\%$ below the direct shares, but
the model ordering is unchanged. Figure~\ref{fig:dilution} is therefore a descriptive cross-check;
the causal evidence for the backbone lever comes from the rescaling intervention.

\end{document}